\documentclass[conference]{IEEEtran}
\IEEEoverridecommandlockouts
\usepackage{cite}
\usepackage{amsmath,amssymb,amsfonts}
\usepackage{url}
\usepackage{algorithmic}
\usepackage{graphicx}
\usepackage{svg}
\usepackage{textcomp}
\usepackage{xcolor}
\usepackage{diagbox}
\def\BibTeX{{\rm B\kern-.05em{\sc i\kern-.025em b}\kern-.08em
    T\kern-.1667em\lower.7ex\hbox{E}\kern-.125emX}}
\begin{document}

\title{Semi-Automated Data Annotation in Multisensor Datasets for Autonomous Vehicle Testing
}


\author{\IEEEauthorblockN{Andrii Gamalii, Daniel Górniak, Robert Nowak,
Bartłomiej Olber, Krystian Radlak, Jakub Winter \\
All authors contributed equally; the author names are listed in alphabetical order.}
\IEEEauthorblockA{Warsaw University of Technology, Warsaw, Poland }}

\maketitle

\begin{abstract}
This report presents the design and implementation of a semi-automated data annotation pipeline developed within the DARTS project, whose goal is to create a large-scale, multimodal dataset of driving scenarios recorded in Polish conditions. Manual annotation of such heterogeneous data is both costly and time-consuming. To address this challenge, the proposed solution adopts a human-in-the-loop approach that combines artificial intelligence with human expertise to reduce annotation cost and duration. The system automatically generates initial annotations, enables iterative model retraining, and incorporates data anonymization and domain adaptation techniques. At its core, the tool relies on 3D object detection algorithms to produce preliminary annotations. Overall, the developed tools and methodology result in substantial time savings while ensuring consistent, high-quality annotations across different sensor modalities. The solution directly supports the DARTS project by accelerating the preparation of large annotated dataset in the project’s standardized format, strengthening the technological base for autonomous vehicle research in Poland.
\end{abstract}


\section{Introduction}
The development of autonomous vehicles (AVs) relies heavily on the availability of large, diverse, and precisely annotated datasets. These datasets form the basis for training and evaluating perception systems responsible for detecting, classifying, and tracking objects in complex traffic environments. Most publicly available AV datasets, such as KITTI\cite{b1}, nuScenes\cite{b2}, or Waymo Open\cite{b3}, reflect driving conditions typical for Western Europe, North America, or Asia, which differ substantially from those found in Poland. Differences in road infrastructure, signage, weather conditions, and driver behavior create the need for a dedicated Polish dataset that can more accurately represent local traffic scenarios.

The DARTS (Database of Autonomous Road Test Scenarios) project addresses this need by developing a comprehensive, multimodal dataset containing synchronized data from a variety of sensors, including lidar, camera, radar, GPS, and IMU. The collected data will serve as a foundation for designing, testing, and evaluating perception algorithms for autonomous and semi-autonomous vehicles operating under Polish road conditions. The resulting dataset will be made available on a non-profit basis to support the development of national competencies in the field of intelligent transportation and to strengthen the domestic research ecosystem in automotive technologies.

One of the main challenges in building large-scale AV datasets is the annotation process — the assignment of semantic and geometric labels to objects and scenes captured by multiple sensors. Manual annotation, especially of 3D data such as lidar point clouds, is both labor-intensive and costly\cite{b4}. Moreover, maintaining consistency and accuracy across annotators and sensor modalities requires extensive quality control and domain expertise. As a result, data annotation often becomes the most time-consuming and expensive stage of dataset production.

This report presents the design and evaluation of a semi-automated data annotation pipeline that integrates artificial intelligence models with human verification to accelerate the preparation of multimodal autonomous driving datasets. The following sections describe the scope of the task, the system architecture, implementation details, and quantitative results demonstrating the efficiency gains achieved.

\section{Scope of the Task}

The scope of this task encompasses the design, development, and evaluation of a semi-automated data annotation pipeline intended to support the creation of a large-scale, multimodal dataset within the DARTS project. The primary goal is to enable efficient, scalable, and high-quality annotation of sensor data collected from test drives under Polish traffic and environmental conditions. The task addresses the challenge of reducing manual annotation effort while maintaining the precision required for training perception systems compliant with SAE Levels 3–5\cite{b5}.

The scope included the integration of AI-based detection models, human-in-the-loop verification, domain adaptation techniques, anonymization modules to ensure compliance with the GDPR regulations, iterative models retraining on new annotated data and a database-driven management system. The task did not include raw data collection or downstream perception model evaluation.

The outcome of this task is a functional annotation framework capable of processing multimodal data at scale, providing measurable time savings compared to fully manual annotation workflows. Additionally, the project introduces a new metric for annotation acceleration, enabling the quantitative evaluation of pipeline efficiency and model-assisted annotation speedup. The tools, methodologies, and data structures developed under this task form a core element of the DARTS infrastructure, directly supporting the preparation of the national autonomous driving dataset in its standardized format.

\section{System Design}

This section provides an overview of the pipeline’s overall structure, workflow, and human-in-the-loop approach, as well as its key software and infrastructure components. It introduces the system architecture at a high level of abstraction that explains how data flows through various processing stages and how automation and human expertise are combined to achieve both scalability and accuracy. 

\subsection{System overview}

\begin{figure}[htbp]
\centering
\includegraphics[width=0.49\textwidth]{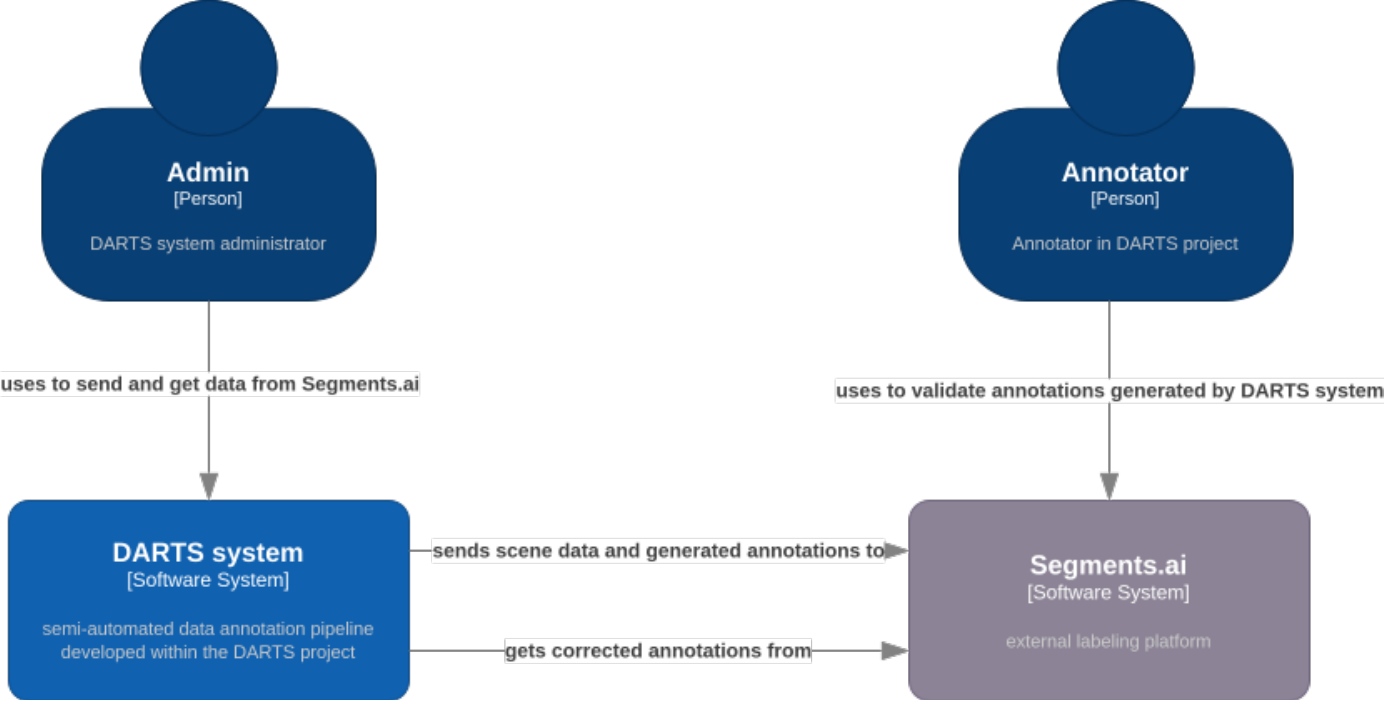}
\caption{System context diagram}
\label{fig:system_context_diagram}
\end{figure}

The general architecture of the semi-automated annotation system is illustrated in Fig.~\ref{fig:system_context_diagram}. The DARTS system is designed as a central coordination layer that connects human annotators, administrative tools, and the external annotation platform Segments.ai.

At the highest level, the Admin interacts directly with the DARTS system to manage datasets, initiate annotation jobs, monitor progress, and retrieve processed results. The DARTS system automates the generation of initial annotations based on sensor data using artificial intelligence models and domain adaptation techniques.

Automatically generated annotations together with scene metadata are sent from the DARTS system to Segments.ai, where they are made available to human annotators. After verification and correction by annotators, the validated annotations are retrieved back into the DARTS system for further processing, quality control, and potential retraining of detection models.

The Annotator interacts exclusively with Segments.ai, reviewing and validating machine-generated annotations. This human-in-the-loop process ensures high accuracy while maintaining efficiency by limiting manual effort to verification and correction rather than full manual labeling.

\begin{figure}[htbp]
\centering
\includegraphics[width=0.49\textwidth]{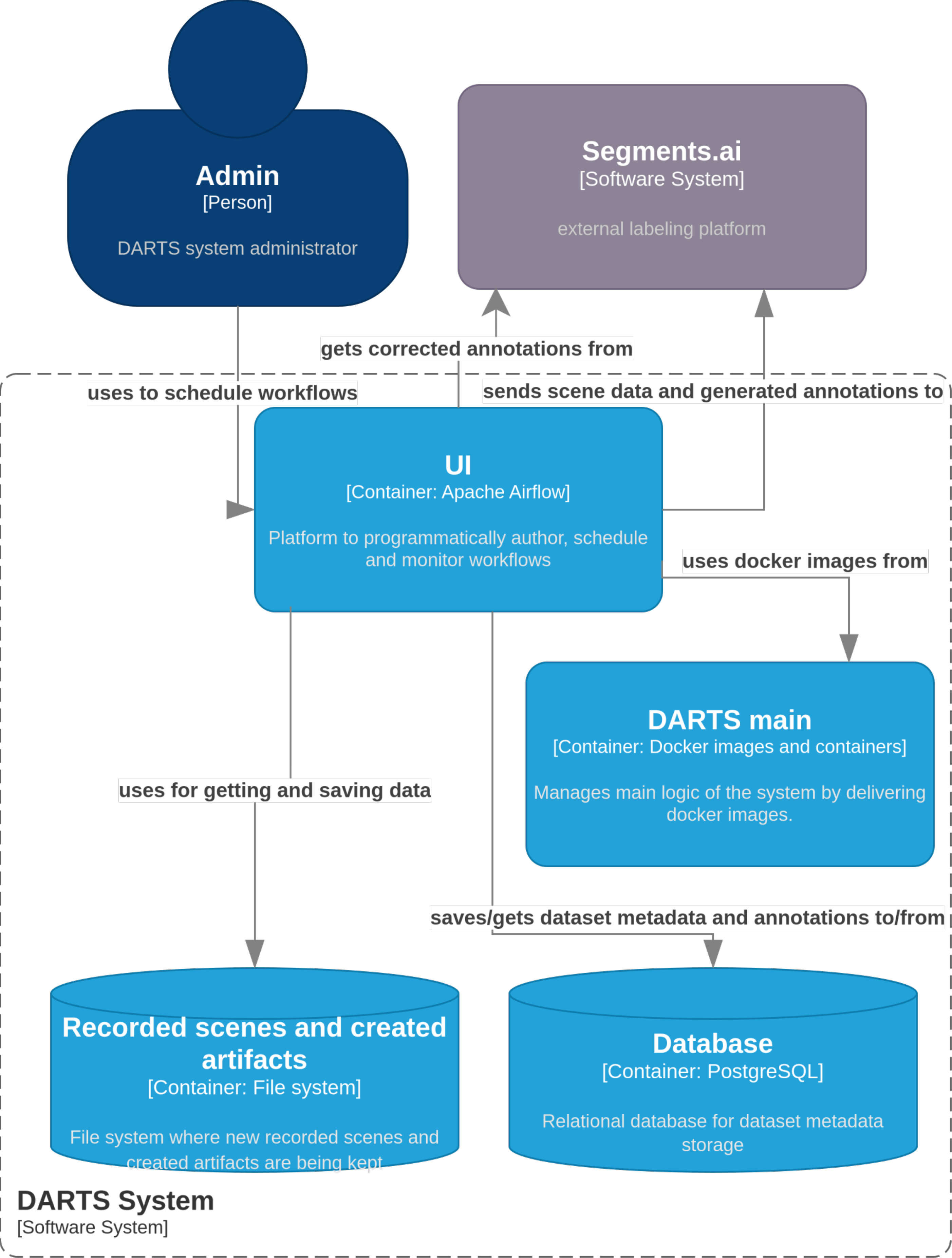}
\caption{Container diagram}
\label{fig:container_diagram}
\end{figure}

The overall architecture of the DARTS system is presented in Fig.~\ref{fig:container_diagram}, which illustrates the primary containers, data stores, and external systems that compose the semi-automated annotation pipeline.

The Administrator interacts with the system through an Apache Airflow instance, which serves as the orchestration platform for managing workflows. These workflows encompass a variety of tasks, such as preprocessing newly recorded raw data into a standardized target format suitable for insertion into the relational database, importing processed data and automatically generated annotations into the Segments.ai platform, as well as retrieving validated annotations and storing them back into the database.

The DARTS main container comprises multiple microservices, each responsible for a specific function within the pipeline — for instance, image anonymization or automated annotation generation. Each service is distributed as a dedicated Docker image, which is executed as part of Airflow-managed tasks, ensuring modularity, scalability, and reproducibility across the system.

Database container is our source of truth in the system. Here we keep metadata about recorded scenes in the system as well as created annotations during the pipeline workflow. Many microservices from DARTS main container use this database to get current information about dataset that is being annotated.

Both the recorded raw data, the preprocessed datasets, and the artifacts produced during workflow execution (such as model checkpoints) are stored in the project’s file system, forming a structured and traceable repository of all data and intermediate results.

\subsection{System workflow and orchestration overview}

\begin{figure*}[htbp]
\centering
\includegraphics[width=0.86\textwidth]{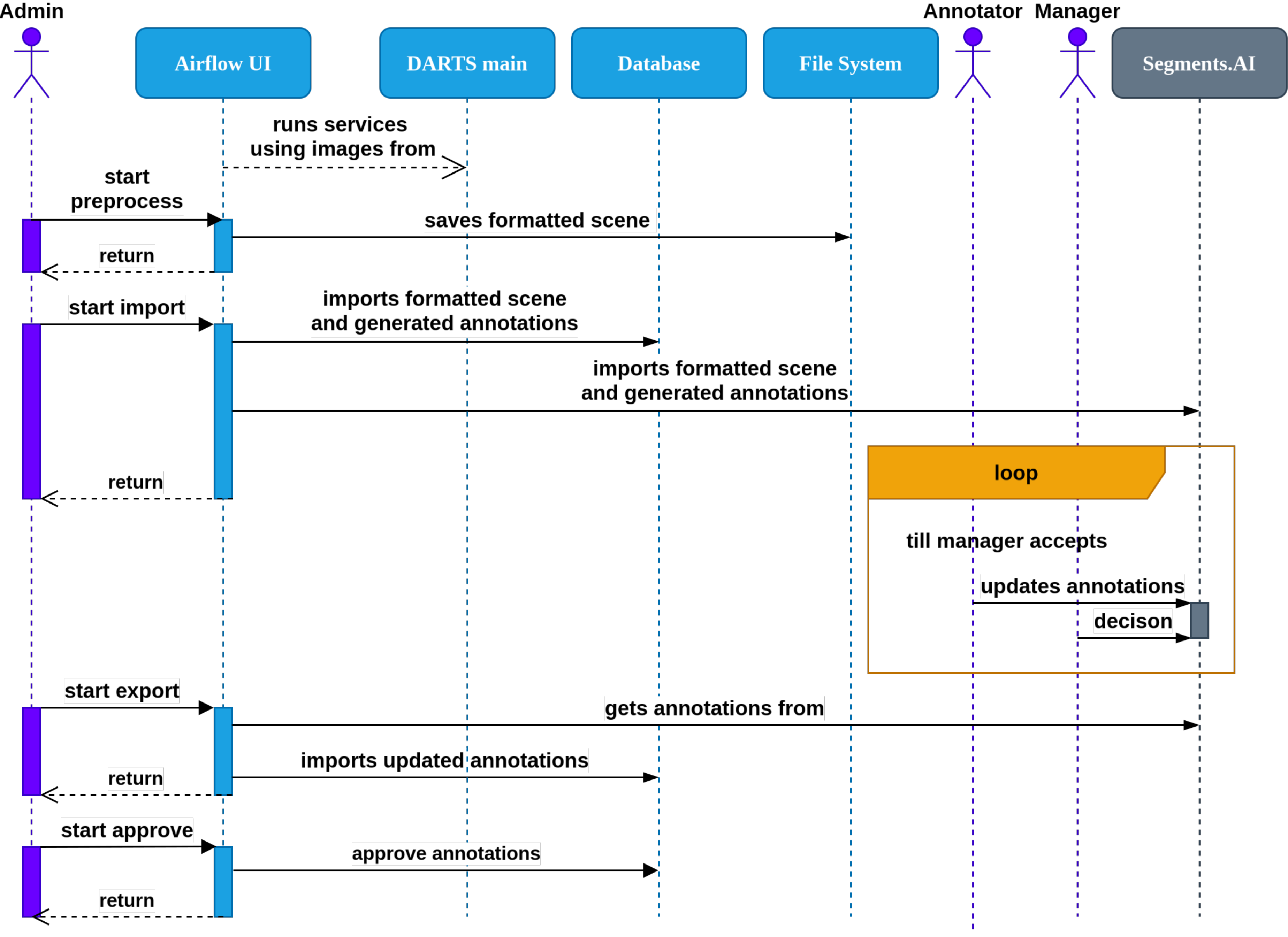}
\caption{Sequence diagram presenting DARTS pipeline}
\label{fig:sequence_diagram}
\end{figure*}

The DARTS annotation pipeline that is shown in Fig.~\ref{fig:sequence_diagram} is orchestrated by an Apache Airflow instance, which manages and monitors stages of the data processing and annotation workflow. Each phase of the system operation is represented as a directed acyclic graph (DAG) that defines the sequence, dependencies, and execution environment of individual tasks. This approach provides high modularity, scalability, and full reproducibility of the annotation process by few clicks in the UI. Airflow tasks use Docker Operator to be able to use services inside docker images from DARTS main container. This allows for decoupling of airflow and microservices development cycles. Airflow serves as the central coordination layer, ensuring that every step is executed in the correct order and that intermediate artifacts are properly logged and stored.

The workflow begins when new scene raw data is recorded during DARTS test drives and uploaded to the system. These raw files, containing lidar, camera, radar, and auxiliary sensor data, are processed using a dedicated Airflow DAG that firstly checks integrity and correctness of data. Then raw data is converted into the standardized DARTS data format with proper calibration and synchronization applied. The resulting data is stored in the project’s file system.

Once data preprocessing is complete, a second Airflow DAG initiates the import and annotation generation phase. This invokes model-based services to automatically generate preliminary annotations for objects detected in camera and lidar data. The system then uploads both the processed data and corresponding annotations to the Segments.ai platform, where they become accessible to human annotators. At the same time, relevant metadata and annotation records are registered in the PostgreSQL database to maintain consistency and traceability.

\begin{figure}[htbp]
\centering
\includegraphics[width=0.49\textwidth]{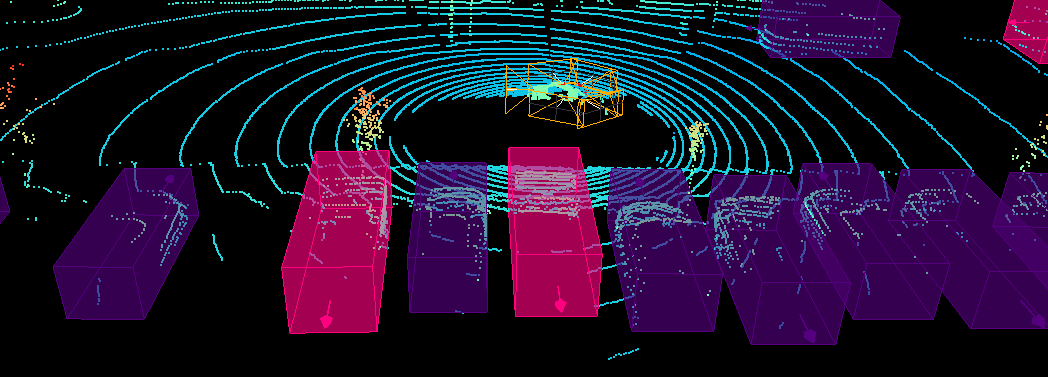}
\caption{Generated annotations imported into Segments.ai}
\label{fig:segmnets_ai}
\end{figure}

In the next phase, human annotators review, refine, and validate the automatically generated annotations within the Segments.ai interface. The annotation manager subsequently performs a final quality control and approval step, ensuring that the resulting labels meet project standards. Once validated, Airflow executes another DAG responsible for importing the corrected annotations into the PostgreSQL database.

The final phase of the workflow involves approval. An Airflow DAG updates the status of the validated annotations to 'approved', finalizing their inclusion in the DARTS dataset. This stage effectively closes the annotation cycle for a given scene and prepares it for potential inclusion in training for subsequent model iterations.

\section{Modules Implementation}

This section presents the implementation details of the individual modules that make up the DARTS semi-automated data annotation pipeline. Each module has been designed as an independent, containerized service responsible for a specific function within the overall workflow — from data preprocessing and AI-based annotation generation to anonymization, synchronization, and database management. Together, these modules form a cohesive and extensible system architecture that supports efficient data flow, modular development, and robust error isolation.

\subsection{Annotation Generator}

The Annotation Generator module is responsible for the automated generation of 3D object annotations from LiDAR point cloud data. It forms one of the core components of the DARTS system, enabling scalable, semi-automated labeling of large datasets while maintaining a high level of accuracy through domain adaptation techniques.

The module builds upon a modified version of the OpenPCDet\cite{openpcdet2020} framework developed by the OpenMMLab community. This framework provides a flexible and modular codebase for 3D object detection using point cloud data, which has been extended in DARTS to support our custom data formats, training routines, and integration with other system components.

For model initialization, the module uses a pre-trained DSVT\cite{WangShi2023} model originally trained on the nuScenes dataset, which provides a strong baseline for general driving environments. To address discrepancies between the nuScenes domain and our in-house sensor configuration, the model was further fine-tuned using data from the Zenseact Open Dataset\cite{zenseact}, chosen for its similar LiDAR sensor setup and environmental characteristics. This domain adaptation step ensures that the model can generalize effectively to our specific deployment conditions without extensive retraining from scratch.

The annotation generation process operates on metadata exported from the PostgreSQL database and the corresponding LiDAR scene data retrieved from the file system. The module ingests these inputs, performs inference using the fine-tuned DSVT model, and outputs detection results in the DARTS JSON annotation format, which is used across the system for consistency and interoperability.

During training or fine-tuning phases, the module produces intermediate and final artifacts, including model checkpoints, performance logs, and configuration files. These are systematically stored in the file system, following a hierarchical folder structure determined by the learning parameters (such as dataset name, number of epochs, or learning rate). This structure facilitates reproducibility, experiment tracking, and model version control within the DARTS ecosystem.

\subsection{Anonymization}

The Anonymization Module is a critical component of the DARTS system responsible for ensuring compliance with privacy and data protection regulations such as RODO. Its primary function is to automatically identify and later obscure personally identifiable information within image data, specifically human faces and vehicle license plates, before any data is shared outside of project.

Face detection is performed using the RetinaFace\cite{deng2019retinaface} model, implemented through the batch-face framework. This model provides high-accuracy detection across a wide range of facial poses and lighting conditions. License plate detection is handled by a YOLO model. To improve its accuracy and robustness in real-world driving scenes, the YOLO model was further fine-tuned on the publicly available dataset from Roboflow Universe – License Plate Recognition.

The module outputs results in the DARTS JSON annotation format, providing 2D bounding box annotations for both faces and license plates. These annotations can either be stored as metadata for downstream use or applied directly to anonymize the images. When provided with both image data and annotation files, the module applies a Gaussian blur to the detected regions, effectively anonymizing sensitive content while maintaining overall image context.

This dual functionality—automatic annotation and optional anonymization—makes the module highly flexible. It can operate as a preprocessing step in dataset generation or as an on-demand anonymization service integrated into the main DARTS workflow. By automating the detection and obfuscation of sensitive elements, the module reduces manual workload for annotators while ensuring that data privacy requirements are consistently met.

\subsection{Data Preprocessing}

The Data Preprocessing Module serves as the entry point for newly collected sensor data within the DARTS pipeline. Its main objective is to transform raw multimodal sensor recordings into the standardized DARTS dataset format, which extends the widely adopted nuScenes format. This ensures that all data entering the system follows a unified structure that supports efficient downstream processing, annotation, and model training.

The preprocessing pipeline performs a comprehensive series of validation checks to guarantee the integrity and usability of the incoming data. These checks include verifying the existence of all required files, ensuring temporal alignment and proper timestamp synchronization across sensors, and confirming overall data consistency through checksum verification and log analysis.

When inconsistencies or validation errors are detected, the module generates a PDF report summarizing all failed checks, providing detailed diagnostic information for further inspection. This report is stored in the project’s designated artifact directory and distributed via email notifications to system administrators.

By enforcing data quality at the earliest stage of the pipeline, this module plays a crucial role in maintaining the reliability and consistency of the entire DARTS ecosystem.

\subsection{Database}
The Database Module forms the central backbone of the DARTS data management architecture and serves as the single source of truth for the entire annotation pipeline. It is responsible for storing, organizing, and maintaining all dataset-related information, including sensor metadata, annotation versions, and dataset lineage. The database acts as the key integration point between the automated components of the DARTS system and the external Segments.ai platform.

The schema implemented in this module is a modified and extended version of the nuScenes dataset format, adapted to the specific needs of the DARTS project. In addition to standard nuScenes entities such as scenes, samples, and sensor data, the schema incorporates support for 2D annotations, the origin and validation status of annotations and scene metadata such as weather during recording. This allows the system to precisely track whether annotations were automatically generated, manually corrected, or approved through human-in-the-loop workflows.

To ensure flexibility and maintainability over time, the Database Module includes a schema migration mechanism alembic that allows the structure of the database to evolve alongside the system. This mechanism enables safe, incremental modifications, such as adding new tables, fields, or relationships, without data loss or interruption to ongoing workflows. As a result, the database can accommodate new data modalities, annotation types, or workflow stages as the DARTS project continues to expand.

\subsection{MOT Tracking}

The Multiple Object Tracking (MOT) Module is responsible for maintaining object continuity across time, ensuring that detections from consecutive frames are correctly associated with persistent real-world entities. While object detectors typically treat each detection as an independent instance, in real-world driving scenarios it is essential to track objects across sequential frames to achieve consistent object identifiers and trajectories. This capability is crucial for training perception systems that must reason about dynamic interactions in time, such as vehicle motion, pedestrian movement, or stationary obstacle detection.

The module is based on an adapted version of the open-source 3D Multi-Object Tracker implementation\cite{mot}. Several modifications were introduced to better align the tracker with the requirements of the DARTS dataset format and to improve tracking robustness in multimodal driving scenarios.

First, the outputs of the Kalman filter are utilized directly to produce updated object annotations, enabling the tracker to refine the estimated object positions between frames. Second, the module incorporates mechanisms for size consistency correction, ensuring that the physical dimensions of tracked objects remain stable throughout their trajectories. Third the detection of stationary objects was added, which is often ignored by standard MOT pipelines, and the approximation of object trajectories using polynomial fitting to smooth and regularize motion paths over time.

The MOT module operates within the DARTS annotation pipeline as a post-processing step applied to automatically generated object detections. It takes as input annotations in the DARTS JSON format, which represent frame-wise detections from the object detector, and outputs a refined annotation set in the same format, enriched with consistent instance identifiers.

\subsection{Segments Toolkit}

The Segments Toolkit Module provides the integration layer between the DARTS system and the external Segments.ai annotation platform. Its primary purpose is to enable automated and reliable synchronization of datasets and annotations between the internal DARTS infrastructure and the collaborative annotation environment used by human annotators.

The module automates the process of creating new datasets on Segments.ai based on information retrieved from the PostgreSQL database and metadata generated within the DARTS pipeline. This reduces manual setup time.

It also handles the import into Segments.ai automatically generated annotations with metadata describing their origin, including which DARTS service or model produced them (for example, the 3D annotation generator or the anonymization module).

After the human-in-the-loop validation and approval steps are completed on Segments.ai, the Segments Toolkit Module is responsible for exporting the reviewed annotations back into the DARTS system. During this process, it preserves information about who corrected annotations, ensuring full accountability and traceability. The exported data are stored in the DARTS PostgreSQL database.

\subsection{Darts Utils}

The DARTS Utils Module serves as a general-purpose toolkit supporting a wide range of operations on annotations throughout the DARTS data annotation pipeline. It provides a unified interface for processing, analyzing, visualizing, and evaluating annotation data produced by both automated models and human annotators.

A key responsibility of the DARTS Utils Module is maintaining a set of core model classes used across the entire pipeline. These include, among others, the DARTSAnnotations and DARTSAnnotations2D classes, which define the standardized internal structure of 3D and 2D annotations, respectively, and the PipelineMetaData class, which stores information about the origin of annotations present in file.

Another major functionality of the DARTS Utils Module is to analyze differences between annotation versions, particularly between the model-generated annotations imported into Segments.ai and those corrected by human annotators so we are aware how much they had to be changed.

The module further supports visualization and inspection of annotations by integrating with the Rerun open-source visualization framework\cite{rerun}. This integration enables developers and researchers to interactively explore lidar point clouds, bounding boxes, trajectories, and other annotation types without sending them to Segments.ai.

For quantitative evaluation, the DARTS Utils Module provides tools for assessing both annotation quality and efficiency. When ground-truth labels are available, it performs accuracy evaluation using the Average Precision (AP) metric adapted from the NuScenes evaluation code. Additionally, the module introduces a custom Correction Acceleration Ratio (CAR) metric developed within the DARTS project. CAR measures the relative time savings achieved by using model-generated annotations as a starting point compared to manual annotation from scratch, providing a direct measure of the practical impact of automation on annotator productivity.
\section{Evaluation and Results}

Traditional performance metrics provide insight into model detection accuracy but fail to capture their practical usefulness in the context of data annotation. To address this gap, a new performance metric called the Correction Acceleration Ratio (CAR) was introduced. The following subsections describe the motivation for developing this metric, its formal definition, and its interpretation in the context of the DARTS annotation pipeline.

\subsection{Motivation for a New Metric}

Widely used 3D object detection metrics quantify a model’s detection performance but do not directly reflect the effort required by human annotators to correct the model’s outputs. For example, a false negative (a missed detection) may require substantially more time to correct than a false positive (an unnecessary detection). As a result, traditional perception metrics may favor models that appear accurate statistically but do not necessarily minimize annotation time.

In the context of the DARTS project, where the annotation pipeline is designed as a human-in-the-loop system, this limitation becomes particularly significant. The focus is not only on how accurately the model detects objects, but on how much it accelerates the overall labeling process when its outputs are verified and refined by human experts.

CAR metric was therefore developed to evaluate a model’s usefulness in practical annotation scenarios. It quantifies the time savings achieved by using automated pre-annotations and enables comparison of various 3D detectors based on their contribution to annotation speed.

\subsection{Metric Definition}

The CAR metric formalizes the annotation process as a combination of automated pre-annotation and manual correction. Each model-generated annotation can introduce one or more errors, and each error type requires a different average time to correct.

A model may commit six simple error types which are false positive, false negative, translation error, rotation error, scale error and classification error. Because positional errors (translation, rotation, scale) may co-occur, four complex error types are also defined to reflect more realistic correction scenarios. In annotation tools, translating, rotating, and scaling are often performed jointly, so their combined correction time is not strictly additive. Those are translation with rotation error, rotation with scale error, translation with scale error and translation with rotation with scale error.

The \textbf{correction time} $C$ represents the total estimated time required for annotators to fix all pre-annotations until they match the ground truth:
\begin{equation}
    C = \sum_{e \in E} t_e n_e, \text{where:}
    \label{eq:correction_time}
\end{equation}
\begin{itemize}
    \item $E$ – the set of all error types,
    \item $t_e$ – the average time to correct an error of type $e$,
    \item $n_e$ – the number of errors of that type.
\end{itemize}

To compare this to a fully manual annotation process, we define the \textbf{baseline time} $B$ as the total time required to create all annotations from scratch:
\begin{equation}
    B = t n
    \label{eq:baseline_time}
\end{equation}
where $t$ is the average time to create one annotation and $n$ is the total number of ground-truth objects.

Finally, the CAR metric is defined as:
\begin{equation}
    CAR = 1 - \frac{C}{B}
    \label{eq:car_definition}
\end{equation}

This formulation provides a direct measure of annotation acceleration. For example a value of 1 indicates perfect automation and value of 0 indicates the model provides no benefit over manual annotation. On the other hand values below 0 indicate that model errors slow down the process.

\subsection{Results}

To estimate correction times for each error type, we conducted a manual study using modified annotations from the nuScenes dataset. Frames were loaded into the Supervisely annotation tool, and trained annotators measured the time required to fix deliberately introduced errors. Measurements were collected for all error types across three object classes (Car, Pedestrian, Cyclist) and two distance ranges.

The averaged results showed that \textbf{false negatives} required the longest correction time (about 23~s), as they involve creating new annotations, while \textbf{false positives} and \textbf{classification errors} were the fastest to correct (about 1–2~s). Positional errors such as translation, rotation, and scale required intermediate times (5–16~s). Complex errors confirmed non-additive correction behavior, supporting their inclusion in the CAR metric.

Threshold values for translation, rotation, and scale errors were established using manually annotated nuScenes sequences in two tools for manual lidar data annotation (Supervisely and Scalabel).  Thresholds were defined such that 90\% of manual annotations fell within them.




To evaluate the performance of various state-of-the-art 3D object detectors used to automatically generate object annotations, we conducted experiments on four publicly available datasets: Zenseact\cite{zenseact}, nuScenes\cite{b2}, Waymo\cite{b3}, and KITTI\cite{b1}. We measured their performance using Average Precision (AP) and introduced the CAR metric for further analysis. The results are summarized in Tables \ref{tab:ap} and \ref{tab:car}.

As shown, the performance of 3D object detectors varies across datasets depending on the chosen metric. Overall, the DSVT detector achieves the best performance on most of the datasets, and therefore DSVT was used in our pipeline.
 
\begin{table}[h!]
\centering
\caption{AP metric of various 3D object detectors across different datasets for the car/vehicle class\label{tab:ap}}
\label{tab:car_models}
\begin{tabular}{lcccc}
\hline
{\diagbox{\textbf{Model}}{\textbf{Dataset}}} & \textbf{Zenseact} & \textbf{nuScenes} & \textbf{Waymo} & \textbf{KITTI} \\
\hline
PointPillars \cite{LangVora2019}& 0.82 & 0.81 & 0.70 & 0.78 \\
PointRCNN\cite{ShiWang2019} & 0.38 & 0.44 & 0.51 & 0.81 \\
PV-RCNN++\cite{shi2023pvrcnnpp} & 0.92 & 0.80 & 0.77 & \textbf{0.84} \\
DSVT\cite{WangShi2023} & \textbf{0.94} & \textbf{0.87} & \textbf{0.78} & 0.76 \\
Second\cite{YanMao2018} & 0.92 & 0.82 & 0.71 & 0.82 \\
CenterPoint\cite{YinZhou2021} & 0.93 & 0.85 & 0.73 & 0.80 \\
TransFusion-L\cite{BaiHu2022} & 0.93 & \textbf{0.87} & 0.74 & 0.74 \\
\hline
\end{tabular}

\begin{flushleft}
\footnotesize{Note: AP values are reported following the evaluation protocols of each benchmark, except Zenseact in case of which the nuScenes evaluation protocol was used. The official validation subset of each dataset was used. Bold values indicate the best performance per dataset.}
\end{flushleft}

\end{table}

\begin{table}[h!]
\centering
\caption{CAR metric of various 3D object detectors across different datasets for car/vehicle class}
\label{tab:car}
\begin{tabular}{lcccc}
\hline
{\diagbox{\textbf{Model}}{\textbf{Dataset}}} & \textbf{Zenseact} & \textbf{nuScenes} & \textbf{KITTI} \\
\hline
PointPillars & 0.72 & 0.62 & 0.83 \\
PointRCNN & 0.49 & 0.51 & 0.82 \\
PV-RCNN++ & 0.92 & 0.84 & \textbf{0.85} \\
DSVT & \textbf{0.93} & \textbf{0.87} & 0.83 \\
Second & 0.89 & 0.63 & 0.84 \\
CenterPoint & 0.91 & 0.76 & \textbf{0.85} \\
TransFusion-L & 0.91 & 0.83 & 0.80 \\
\hline
\end{tabular}

\begin{flushleft}
\footnotesize{Note: CAR values are reported for the following distance ranges: [0, 50]m for nuScenes and Zenseact and [20, 50]m for KITTI. The official validation subset of each dataset was used. Bold values indicate the best performance per dataset.}
\end{flushleft}

\end{table}

\subsection{Summary}

The conducted experiments demonstrate that the proposed CAR metric is an effective and intuitive metric for assessing the real benefit of automated annotation in human-in-the-loop systems such as DARTS. Unlike classical perception metrics (e.g., AP), CAR directly quantifies the time savings achieved through pre-annotation and correction efficiency, offering a more practical measure of model utility in dataset creation workflows.

Empirical results demonstrate that integrating 3D object detectors into the DARTS pipeline significantly reduces manual annotation time. On the Zenseact dataset, which closely resembles the data collected in the DARTS project, the highest observed CAR value of 0.93 demonstrates a substantial reduction in manual annotation effort. Specifically, this result indicates that annotation time can be reduced by up to 93\% compared to a fully manual process. Consequently, we can expect a similar reduction in annotation effort for DARTS data relative to fully manual annotation.




\section*{Acknowledgment}
This work was supported by the National Centre for Research and Development under the research project \mbox{GOSPOSTRATEG8/0001/2022}.

\bibliographystyle{IEEEtran}
\bibliography{bibliography}

\end{document}